# FPGA-based Binocular Image Feature Extraction and Matching System

Qi Ni
Harbin Institute of Technology, Shenzhen

Fei Wang
Harbin Institute of Technology, Shenzhen

Ziwei Zhao
Harbin Institute of Technology, Shenzhen

Peng Gao
Harbin Institute of Technology, Shenzhen

## ABSTRACT
Image feature extraction and matching is a fundamental but computation intensive task in machine vision. This paper proposes a novel FPGA-based embedded system to accelerate feature extraction and matching. It implements SURF feature point detection and BRIEF feature descriptor construction and matching. For binocular stereo vision, feature matching includes both tracking matching and stereo matching, which simultaneously provide feature point correspondences and parallax information. Our system is evaluated on a ZYNQ XC7Z045 FPGA. The result demonstrates that it can process binocular video data at a high frame rate (640 × 480 @ 162fps). Moreover, an extensive test proves our system has robustness for image compression, blurring and illumination.

## CCS Concepts
•Hardware ➝ Integrated circuits ➝ Reconfigurable logic and FPGAs ➝ Hardware accelerators

## 1. INTRODUCTION

Machine vision has been widely used in industrial testing, autonomous driving even in consumer electronics. Feature extraction and feature matching are pre-stage algorithms for many applications, such as target tracking, image stitching, visual odometer, SLAM, AR and VR. Feature extraction includes both feature detection and feature description. As long as a feature point is detected, a descriptor (generally a vector) is built up using information around it. If the distance between a pair of descriptors is small enough, their corresponding feature points are considered "matched". After feature matching, all possible corresponding point pairs between two images will be found, and then holography between the two images can be defined. Repeatability is a metric for evaluating the performance of a kind of specific defined feature. It reflects the robustness of a specific kind of feature against image changes caused by camera movement, sensor noise, illumination, and loss compression. SIFT[1]\SURF[2] are two well-known feature extraction algorithms, due to their excellent invariance in rotation, affine transformation, illumination variation. SURF[2] algorithm, as a speedup version of SIFT, takes advantage of integral image and box filters to reduce complexity while with similar accuracy. However, SURF are still very time-consuming. Moreover, descriptors of SIFT and SURF are both complex and need huge memory overhead. A BRIEF descriptor [3], in a binary vector form, is fast both to build and match. Other feature definition may include Harris[4], FAST[5], ORB[6], etc.

Feature extraction and matching are computation intensive, which are resource starvation for storage and memory bandwidth. For instance, it takes 717ms, 180ms and 755ms on SURF detection, BRIEF descriptor building and BRIEF matching respectively on a desktop CPU (E8400 @ 3GHz @512 × 384) [7]. Therefore, parallel versions of SURF\SIFT attract wide attention from researchers. A GPU (GTX480) accelerator for SURF detection and description (no matching) reaches 40FPS @791 × 740[8]. Another instance of GPU (GeForce 8800M) accelerator shows SURF feature matching (1024 feature points) takes only 19ms[9]. However GPU acceleration approach is very power consuming. Other acceleration approaches are based on FPGA or ASIC have been proposed [10-13]. The design that implements SURF detection and description achieves a speed performance of 72FPS on Stratix III @1080P[12]. In [14], a complete hardware acceleration system which include FAST feature point detection, BRIEF description and matching reaches 308FPS on Zynq @640×480.But robustness of FAST feature points is weaker than that of SURF. Moreover, all of the above works are only for monocular applications. Major contributions of this paper are as follows:

1) A binocular system on FPGA is built, which covers SURF detection, BRIEF description and matching, and runs at a frame rate of 162fps @640×480.

2) Standard AXI protocol is used, and different IPs can be flexibly exchanged or configured. Consequently, the design can be extended to more complex applications.

## 2. FEATURE EXTRACTION

SURF description is too complicated to implement. Consequently, the BRIEF description is used as an alternative. In this paper, we choose the combination of SURF feature detection and BRIEF description, whose rationality is discussed in [3]. Of cause this combination sacrifice rotation invariance and partial scale invariance. But in many applications, robotic, like quad-drones, may move smoothly, where requirement for rotation invariance may not be indispensable. To help readers to understand our circuit design, SURF detection and BRIEF description are reviewed shortly.

### 2.1 SURF Detection

SURF consists of three main steps: image integration, calculation of Hessian determinant and non-maximum suppression.

Image integration is one of the key innovations in SURF, which greatly increases the speed of the entire algorithm. It is calculated as

$$I_\Sigma(x,y) = \sum_{i=0}^{x}\sum_{j=0}^{y} I(i,j) \qquad (1)$$

where $I$ is the pixel in original image, and $I_\Sigma$ is the pixel in integral image.

In the SURF algorithm, pixels with higher Hessian determinant values are considered feature points. Hessian matrix is

$$H(x,y,s) = \begin{bmatrix} D_{xx}(x,y,s) & D_{xy}(x,y,s) \\ D_{xy}(x,y,s) & D_{yy}(x,y,s) \end{bmatrix} \qquad (2)$$

where $H(x,y,s)$ is the Hessian matrix of scale $s$. In SURF, a box filter is used to approximate a Gaussian filter. Thus $D_{xx}$, $D_{xy}$ and $D_{yy}$ represents an approximated second-order Gaussian derivative. The Hessian determinant can be calculated by

$$\det(H) = D_{xx}D_{yy} - (\omega D_{xy})^2 \qquad (3)$$

The coefficient $\omega$ is used to correct the error caused by approximation.

In the original SURF algorithm[1], the scale space was established for the purpose of scale invariance. The space is divided into multi octaves, and each of them has 4 scales. To save logic resource, we only construct an octave that has the smallest 8 scales ($s = 1.2, 2.0, 2.8, 3.6, 4.4, 5.2, 6.0, 6.4$), and omit the interpolation step. The rationality of this method has been discussed in [10]

Non-maximum value suppression (NMS) localizes candidate points with local maximum Hessian determinants to prevent feature points from being too concentrated. The NMS search scope includes not only the 8 neighborhoods at its own scale but also the 18 neighborhoods at the two adjacent scales. If the Hessian determinant of the candidate point exceeds a threshold, then the point will be considered as a feature point.

## 2.2 BRIEF Description

To create a BRIEF descriptor, two steps are needed: smooth filtering and binary descriptor generation. In the smooth filtering step, to reduce the number of multipliers, a $9 \times 9$ average filter is used instead of the Gaussian filter recommended in [3]. In the binary descriptor generation step, on the filtered image a $N \times N$ window centered at the feature point is selected, and then 128\256 pairs of the sample points in the window are compared. The comparing result is calculated by

$$\tau(p_1, p_2) = \begin{cases} 1 & if\ (p_1 > p_2) \\ 0 & otherwise \end{cases} \qquad (4)$$

where $p_1$ and $p_2$ are the pixel values of the sample points. All $\tau$ will be spelled into a $M$ bit vector, i.e., the so-called descriptor. In this paper, $N = 49$ and $M=128$.

## 3. THE PROPOSED ARCHITECTURE

### 3.1 Overview

Modern Xilinx Zynq SOC has Processor System and Programmable Logic, as shown in Figure 1. Video streams from two camera are fed into *Image Capture* module. Streams in DVP format are converted to satisfy AXI4 protocol, and then enter a ping-pong buffer in DDR. *Image Rectification* model, similar to the work in [15], extracts stream data form the ping-pong buffer for epipolar rectification and distortion removal. The *Feature Extractor* implements feature detection and descriptor generation, and outputs pixel coordinates and descriptors of feature points. For stereo vision system, *Feature Matcher* not only implements trace matching but also stereo matching. In this paper the left image is used as the reference. The trace matching builds feature point correspondences for left image of current frame and the left image of the previous frame, and the stereo matching builds the correspondences for the left and right image in current frame. The parallax of feature points can be obtained by simple coordinate subtraction according to stereo matching results. Trace and stereo matching results are temporarily stored into DDR by *DMA*. Finally, *ARM APU* transmits those results to a PC through *Ethernet*.

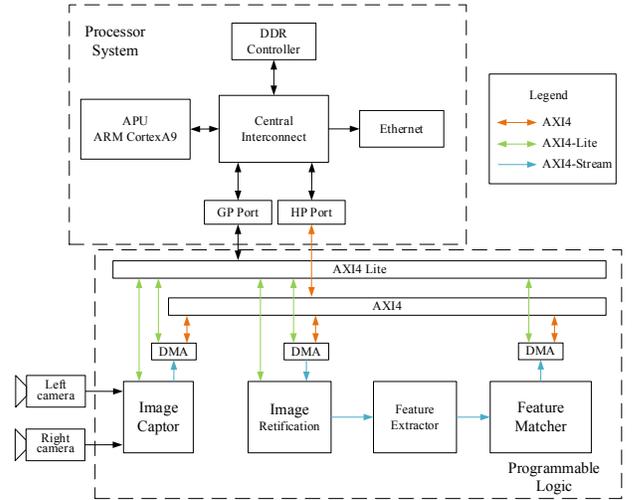

**Figure 1. Block diagram of the proposed system**

### 3.2 Feature Extractor

As a key component, a generic pipelined image filter architecture is introduced as shown in Figure 2. It includes a *Window* module and *Function* modules. Progressive Image data is fed into the *Window*. Except for the last one, each line of Window can hold exactly one line of the image data. In the window, only pixels values in the *Reg* modules can be accessed by the *Function* modules. The *Shift Register* is built with block RAMs and FIFOs. The combination of *Window* and *Function* supports to implement various linear or nonlinear image filter functions.

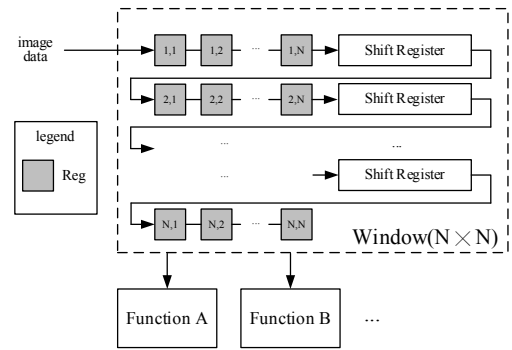

**Figure 2. Architecture of image filter**

The *feature extractor* module is shown in figure 3. It has three types of *Window* (N × N) module in Figure 2, with N equals to 52, 3 and 49 respectively. Image integration, the first step in SURF detection, is implemented by the *Integrator* module. For details of the *Integrator*, please refer to [16].

*Hessian Core* obtains integral image data from *Window* (52 × 52) and calculates Hessian determinant according to Eq (3). There are 8 *Hessian Cores* running in parallel, with each them corresponding to one of the eight scales The Hessian determinant of each pixel on the integral image is first judged by the *Non-Maximal Suppression module* to determine whether it is a local maximum. *Non-Maximal Suppression module* picks those points with local maximum Hessian determinants as candidate points. The candidate points are then fed into *Threshold* module. Only those candidate points whose Hessian determinant is great than the threshold can be thought as a SURF feature point, and a one-bit flag is set as true. The latency forms the input image data to the key point flag is longer than that of the descriptor. Consequently, a FIFO is necessary for data synchronizing. Because the key point flag has only 1 bit, the FIFO consumes little resource.

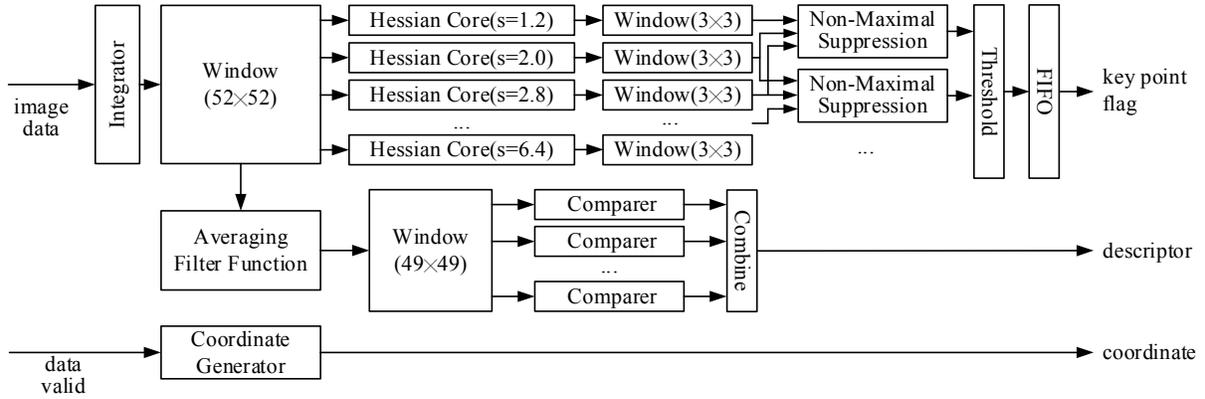

**Figure 3. Block diagram of Feature Extractor**

In BRIEF description, the Gaussian filter is replaced by an averaging filter as mentioned in section 2.1. Unlike Gaussian filter, averaging filter can be implemented by accessing the data of integral image, which means that the *Averaging Filter Function* module can share data in the *Window* (52 × 52) with *Hessian Cores*. Thus, additional *Window* for *Averaging Filter Function* is not needed. Moreover, only one adder and two subtractors need to be used in the *Averaging Filter Function* because of the integral image. The *Compare* module with 1bit output is used to implement Eq (4). Outputs from the 128 *Compare* modules are combined into a BRIEF descriptor. The *Coordinate Generator* module contains two counters for row and column coordinate counting respectively. The input and output of the *feature extractor* are designed to adapt to the AXI-Stream protocol (not shown in Figure 3). The AXI-Stream has a handshaking mechanism. When the input data is invalid or the subsequent circuit is not ready for data reception, the *feature extractor* will be suspended.

### 3.3 Feature Matcher

#### 3.3.1 Multi Buffer module

As depicted in Figure 4, our system is designed as a two-stage pipeline to speedup processing. The matching occurs one frame later than exaction. In one binocular system，feature exaction is conducted both on the left and right images, and trace matching followed by stereo matching has to be conducted The extraction results of current left, current right and previous left frame will be stored for future trace and stereo matching. For example, the 3T (current frame Trace matching) needs the results of 2L (previous left) and 3L (current left). The 3S (current frame Stereo matching) needs the output data of 3L (current left) and 3R (current right) extraction. Note that trace matching only starts from the 2nd frame.

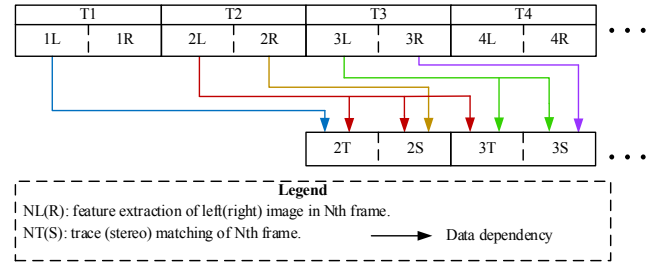

**Figure 4. Schedule of feature exaction and matching**

Extraction and matching are performed simultaneously, i.e., extraction results are read and written at the same time. To avoid conflicts between reading and writing, a Multi Buffer module is proposed. As shown in Figure 5, the Multi Buffer is logically a ring buffer that consist of five storage sections. In any time moment, three sections are read and two are written. For example, at the T3 time, the extraction results of 1L, 2L and 2R are read from RP, RL and RR (Note the legend in Figure 5) sections to perform 2T and 2S matching; at the same, the 3L, 3R results are wrote into WL, WR sections for future matching at T4 and T5.

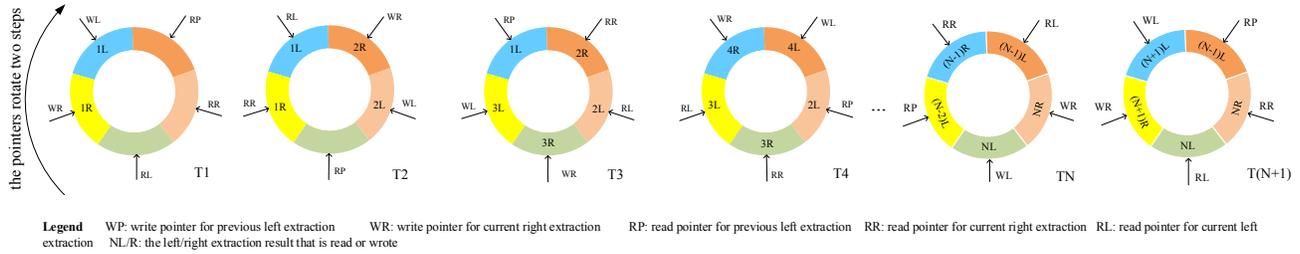

**Figure 5. Principle of *multi buffer***

### 3.3.2 Match Executor module

Figure 6 shows the block diagram of the *Match Executor* module. The inputs of the *Match Executor* module are from the RP, RL and RR storages in the *Multi Buffer* module. The *Match Core* models in the *Multi Buffer* are divided into T group and S group, which are used for tracking match and stereo match respectively. The number of the *Match Core* models can be flexibly configured. The more matching cores, the faster the matching, but the more resources are consumed. The matching process is controlled by a finite state machine (FSM) that will be discussed below.

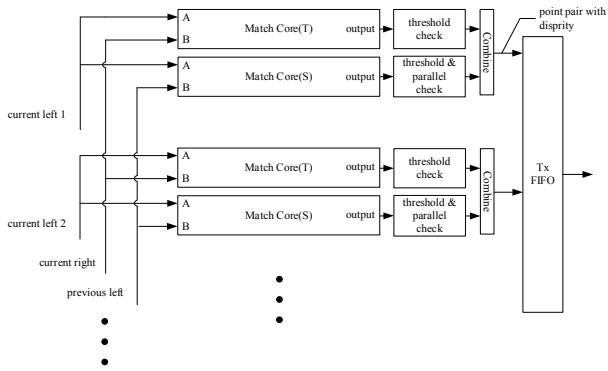

**Figure 6. Block diagram of the Match Executor**

The *Match Core*, as shown in Figure 7, is used to find a pair of descriptors with the smallest Hamming distance [3], and therefore their corresponding feature points. A and B, the inputs of *Match Core* are 148-bit wide (128bit for descriptor and 20bit for coordinate). In matching, the descriptor and the coordinate of a feature point (denoted as FA) will be loaded into A. Only the descriptor portion is used for calculating the Hamming distance, the coordinate portion just accompanies the descriptor portion without any change. Then a group of feature points (denoted as FBs) will be loaded into B respectively. In the process of loading FBs, if the Hamming distance between new FB with the FA is less than that of old FB, the comparator of *Match Core* will output 0 and the *Hamming Distance Register* will be updated. After the process, the best point in the FBs that matches the FA will be found. The "best" means the minimum Hamming distance.

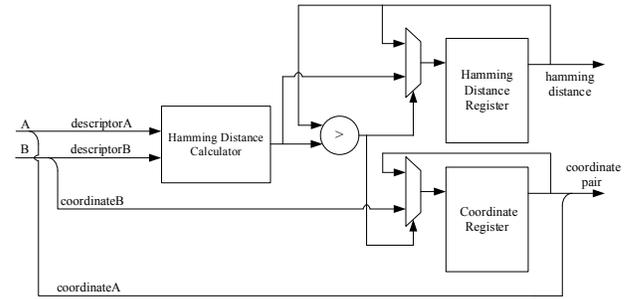

**Figure 7. Block diagram of Match Core**

As depicted in Figure 8, the matching processing is controlled by an FSM, which has four states: LOAD, RUNNING, TRANSPORT and CLEAR. Now matching at time T3 is used as an example for illustration. We assume that there are N group of *Match Cores* in the *Match Executor* Note that the left image is used as the reference image. At LOAD state, there will be N descriptors and coordinates from the RL storage to be loaded into the inputs including current left 1, current left 2… current left N. Before the signal RUNNING DONE becomes true, all descriptors and coordinates from the RR (RP) will be loaded into the input current right (previous left) in order, just like the FBs mentioned before. During this time, all *Match Cores* are working. When RUNNING DONE is true, the *Match Cores (T)* will output N coordinate pairs between left image of 2nd frame and left image of 1st frame. The *Match Cores (S)* will output N coordinate pairs between left and right image of 2nd frame. If the Hamming distance is bigger than the threshold, the coordinate pairs will not pass the *threshold check*. Since *Image Rectification* aligns the epipolar lines in parallel, the difference between the vertical coordinates in a corresponding pair should be close to 0. In trace matching, only *threshold check* is required. But in stereo matching, both *threshold check* and *parallel check* are required. At TRANSPORT state, all *Match Cores* stop working, and the matching results, as well as their coordinate pairs, are shifted to the AXI bus via the *Tx FIFO*. At CLEAR state, all *Match Cores* are reset and then the FSM returns to LOAD state. The FSM repeats in this way until descriptors and coordinates from the RL storage are loaded. The number of feature points is not necessarily divisible by N. In this case, some matching cores will be disabled in the last FSM iteration.

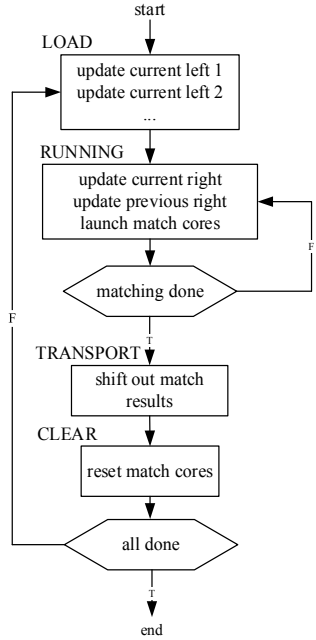

**Figure 8. Flow chart of the FSM that controls matching**

## 4. EXPERIMENTS

### 4.1 Error Analysis

Error is inevitable when floating point data are approximated with fixed-point data, and error analysis is conducted. The $recall \sim 1-precision$ curve [17] is used to evaluate the accuracy of the trace matching，which is defined as:

$$recall = \frac{\#correct\ matches}{\#correspondences} \quad (5)$$

$$1-precision = \frac{\#false\ matches}{\#correct\ matches + \#false\ matches} \quad (6)$$

As shown in Figure 10, performance of the proposed algorithm is compared with some other feature extraction algorithms in OpenCV.

The *Bikes, boat, wall* and *ubc* in the data set correspond to image blur, zoom+rotation, viewpoint change, and JPEG compression, respectively. For the detail of the data set, please refer to [17].Except *boat,* the proposed method performs well on the data set. In the data set *boat,* there is obviously rotation and scale variations between the two images, the proposed method does not performs well. Figure 9 shows the example of trace matching. Mismatched point pairs can be removed by the RANSAC[18]

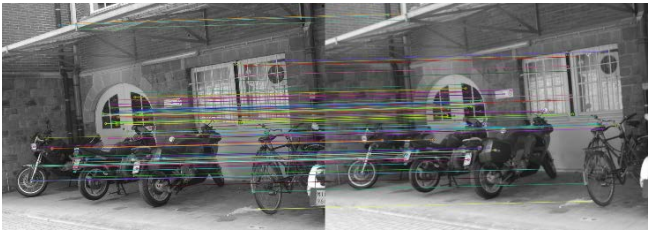

**Figure 9. An example of trace matching result. Images: bikes 1 and 2 in [17]**

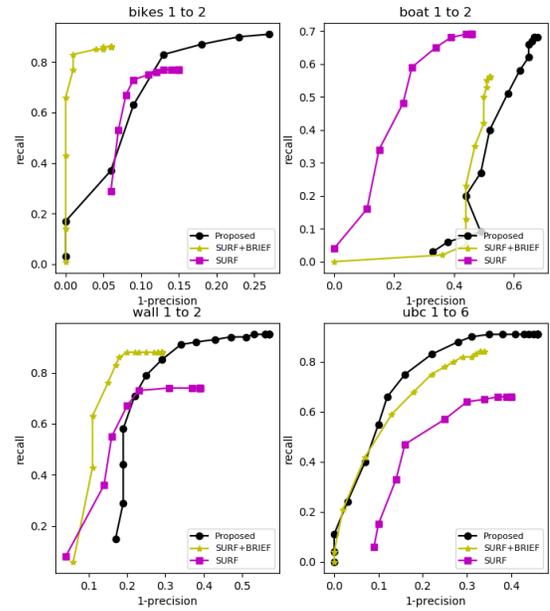

**Figure 10. The comparison between proposed algorithm and other in OpenCV**

Stereo matching can create parallax for a corresponding pair. As mentioned in section 3.3.2, unlike tracking match, stereo matching is restricted by not only the threshold, but also the range of parallax search.

As shown in Figure 10, the *precision* of stereo matching almost reaches 100% for all image pairs in the data set [18], due to the *parallel check* which eliminates point pairs which do not satisfy parallax epipolar line constraint.

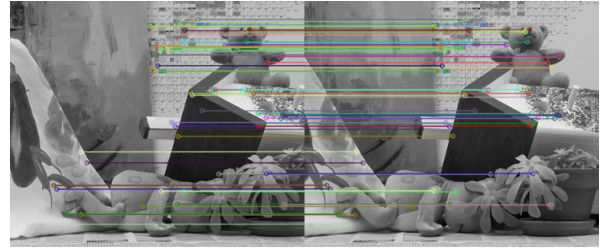

**Figure 11. An example of stereo matching result. Images: teddy 2 and 6 in [19].**

### 4.2 Performance analysis

Assuming an image resolution of $640 \times 480$, the resource utilization on a ZYNQ XC7Z045 FPGA is shown in Table 1.

**Table 1. Utilization on the selected FPGA**

| Module | LUT (as logic) | LUT (as memory) | FF | BRAM | DSP |
|---|---|---|---|---|---|
| detector | 10878 | 5702 | 18435 | 26 | 160 |
| descriptor | 3097 | 2281 | 9272 | 68 | 0 |
| matcher | 3479 | 26 | 2134 | 22.5 | 0 |
| Total | 17454 | 8009 | 29841 | 116.5 | 160 |
| Available | 148200 | 70400 | 437200 | 545 | 900 |

Assuming a 100MHz working frequency, our system can work at 162fps @640 × 480. For contrast, our system and previous works are listed in Table 2. Those works in [14, 20-23] only support monocular cameras. Although our system is targeted to binocular vision, performance metrics of monocular version of our system is also given. For [20, 21], only feature detection and description is accelerated on FPGA, while the matching of them are finished by software. So their frame rates are low Literatures [13, 22, 23] accelerate feature point extraction、descriptor construction and matching. The frame rates of [22, 23] are still low. The performance of [14] is similar to our system in monocular mode because of the similar design on pipeline architecture.

**Table 2. Comparison between our design and previous work**

| Ver. | Det | Des | Arc | Clk | Image Size | FR |
|---|---|---|---|---|---|---|
| [20] | SIFT | SIFT | Stratix+NIOS | 100 | 320×240 | 30 |
| [21] | SURF | BASIS | Virtex+CPU | 400 | 640×480 | 30 |
| [22] | FAST | BRIEF | ZYNQ | 100 | 640×480 | 55 |
| [23] | SIFT | BRIEF | Virtex | 200 | 1280×720 | 60 |
| [14] | FAST | BRIEF | ZYNQ | 100 | 640×480 | 308 |
| Our (MO) | SURF | BRIEF | ZYNQ | 100 | 640×480 | 325 |
| Our (BI) | SURF | BRIEF | ZYNQ | 100 | 640×480 | 162 |

Note: MO: Monocular, BI: Binocular, Det: Detection, Des: Description, Arc: Architecture, Clk: Clock (MHz), FR: Frame Rate (fps)

## 5. CONCLUSION

In this paper, SURF feature detection, BRIEF descriptor construction and matching system is proposed for binocular vision systems. It can work at 162fps @640 × 480 on a ZYNQ XC7Z045. The use of standard AXI4 interfaces allows different modules in the system to be exchanged or configured easily. In future, we will focus on improving the accuracy and combining our system with higher level applications like visual odometer or SLAM.